\documentclass[lettersize,journal]{IEEEtran}
\usepackage{amsmath,amsfonts}
\usepackage{algorithmic}
\usepackage{algorithm}
\usepackage{array}
\usepackage[caption=false,font=normalsize,labelfont=sf,textfont=sf]{subfig}
\usepackage{textcomp}
\usepackage{stfloats}
\usepackage{url}
\usepackage{verbatim}
\usepackage{graphicx}
\usepackage{multirow}
\usepackage{cite}
\begin{document}

\title{Mind the Missing Split: Resolving Feature Heterogeneity in Swarm Learning with Random Forests}

\author{Mohammad Tajabadi\textsuperscript{1} and Dominik Heider\textsuperscript{1}%
\thanks{\textsuperscript{1}Institute of Medical Informatics, University of Münster, Münster, Germany}
}




\maketitle

\begin{abstract}
Swarm Learning is a decentralized collaborative learning mechanism that allows multiple organizations to train a shared model without central coordination or direct data sharing. In typical horizontal Swarm Learning, datasets across sites are usually assumed to share the same feature set. However, in real-world applications, sites often have partially overlapping features because measurements, protocols, and available covariates differ across sites. This feature heterogeneity creates a practical issue for machine learning algorithms such as Random Forests. Specifically, when decision trees are pooled into a global Random Forest, inference at a given site can become ill-defined if a traversal encounters a split on a feature that is not available locally, often forcing organizations to discard site-specific variables upfront. In this paper, we address feature heterogeneity in Swarm Learning with Random Forests under partially overlapping feature spaces. We propose several deterministic and probabilistic inference-time strategies that resolve such missing splits without restricting training to the intersection of features. We evaluate the methods on nine datasets and demonstrate that they outperform both the intersection baseline and locally trained models across a broad range of scenarios.
\end{abstract}


\section{Introduction}
\label{introduction}
\IEEEPARstart{T}{he} rapid growth of Artificial Intelligence (AI) in recent years has encouraged researchers and companies to invest in the development of machine learning models for automation, prediction, and diagnosis in various fields, such as healthcare, education, manufacturing, and finance. For instance, in healthcare, AI models can assist in the discovery of new drugs \cite{dara2022machine}, enhancing diagnosis \cite{Spaenig2019, Ruhland2025}, and treatment \cite{kononenko2001machine, asif2025advancements}. However, the generalizability of machine learning models often depends on the availability of large amounts of data for training \cite{krizhevsky2017imagenet}. Consequently, data sharing and collaborative learning in certain fields have become popular, as they allow organizations to access more data directly or indirectly to build more accurate models. Since data sharing in certain applications, such as healthcare, is challenging due to privacy regulations \cite{tajabadi2023sharing}, techniques such as Federated Learning (FL) \cite{mcmahan2017communication} and Swarm Learning (SL) \cite{warnat2021swarm} have been developed that allow multiple organizations to collaboratively train a model while keeping their private data at their local sites \cite{tajabadi2024fair, torkzadehmahani2022privacy, tajabadi2024privacy}. In FL, the parties form a star network topology in which a central server governs the learning process, while in SL, they form a fully connected network in which they communicate directly with each other without any central server. This peer-to-peer design improves fault tolerance and avoids a single point of control \cite{tajabadi2024privacy}.

The most researched form of FL or SL is the horizontal form, in which the data across sites share the same set of features but differ in samples \cite{yang2019federated}. For instance, two hospitals may have datasets for diabetes prediction with samples from different patients and specific features, such as age, weight, visual blurring, blood sugar, and blood pressure. Conversely, in vertical FL or SL, the data across sites differ in their features but include the same samples. For instance, a hospital and a wearable tech company might collaborate to predict patients' risk of diabetes. In this case, the data in the hospital has features such as weight and blood sugar, and the data on the company's side might feature lifestyle information such as heart rate patterns and sleep quality.

In this article, we aim to address a challenging form of data heterogeneity in horizontal SL in a cross-silo setting, where the data across silos have overlapping and non-overlapping features. We propose various methods with Random Forests (RFs) that enable collaborative learning even with missing features at different sites.

The rest of this paper is organized as follows: In Section \ref{related-work}, we review the related work. In Section \ref{methods}, we introduce our architecture and methods. In Section \ref{experiments-setup}, we introduce our experiments setup, and we present the results in Section \ref{results}. We discuss the implications of the results and future directions in Section \ref{discussion}. Finally, in Section \ref{conclusion} we conclude the paper.

\section{Related Work}
\label{related-work}

The issue of missing data is a classic challenge in real-world applications of machine learning. Numerous methods have been proposed to address this problem. One common strategy is to impute the missing value using single imputation. Simple univariate methods, such as the mean or median imputation, use only the distribution of the feature itself \cite{acuna2004treatment, zhang2016missing}, while more sophisticated multivariable methods use information from other observed features. These include techniques such as regression models \cite{sherwood2013weighted, templ2011iterative} and K-Nearest Neighbour (KNN) models \cite{jiang2015cknni, cheng2014k}. These methods are called single imputation because each missing value is imputed only once. In multiple imputation methods, on the other hand, several complete versions of the dataset are created by imputing multiple values for missing items to account for the uncertainty inherent in an imputation process \cite{rubin2018multiple, kenward2007multiple}. Standard analyses are then performed on each completed dataset, and the results are combined with appropriate methods \cite{graham2007review}. More recently, deep learning techniques, particularly Generative Adversarial Networks (GANs), have been adapted for missing data imputation. For instance, Yoon et al. \cite{yoon2018gain} introduced Generative Adversarial Imputation Nets (GAIN), in which the generator observes components of a data vector, imputes the missing components conditioned on the observed data, and outputs a completed vector. The discriminator’s goal is then to distinguish between observed and imputed components.

Within the domain of tree-based algorithms, specific strategies have been developed to address the issue of missing data. In Classification and Regression Trees (CART), the concept of surrogate splits is proposed to specify replacement splits in case of missing values \cite{breiman2017classification}. The idea is that during training a decision tree, for each primary split on a feature, a list of surrogates, chosen from other features, is calculated based on how closely they mimic the behavior of the primary split. At inference time, if the primary feature's value is missing for a sample, the best available surrogate is used to direct the sample down the tree. The idea of surrogate splits is more commonly used in correlation studies \cite{voges2023exploitation}.


In contrast, the C4.5 algorithm \cite{quinlan2014c4} employs a weighting mechanism for missing values. For a split on feature $F$, the gain ratio is computed using only samples with $F$ observed. After creating the split, if an instance has a known value for $F$, it is assigned to exactly one child with weight 1. If $F$ is missing, the instance is assigned to all children with different weights. Each weight equals the estimated probability that the instance belongs to that child, i.e., the fraction of cases with a known value of $F$ that satisfy that child’s condition at that node. The same fractional assignment is also applied at inference.

More modern gradient boosting implementations like the XGBoost algorithm also incorporate mechanisms for missing data \cite{chen2016xgboost}. XGBoost learns a default route with the maximum gain to direct samples with missing values down a path. During training, it sends all samples with missing values for a specific split to both directions and calculates the gain. Then the path with maximum gain is chosen as the default route. At inference time, any sample arriving at a node with the split feature missing is simply directed down this default path.

In the context of decentralized learning, the issue of data heterogeneity introduces additional complexities. While most research assumes perfectly aligned horizontal or vertical data partitions, real-world scenarios might deviate from this ideal. In this regard, Liu et al. \cite{liu2020secure} introduced Federated Transfer Learning (FTL) to address the heterogeneity problem in federated settings by transferring knowledge from one domain to another, where both the feature-space and the sample-space are different. For instance, Chen et al. \cite{chen2020fedhealth} developed HealthFed, an FTL framework for transferring knowledge in wearable healthcare settings where feature spaces may differ. Federated approaches have also been applied directly to imputation. Zhou et al. \cite{zhou2021federated}, for instance, introduced a federated generative adversarial network (FGAN) model for imputation of air quality data across different sensor nodes.

While the aforementioned methods address classical missing data, specific tree-based heuristics, or transfer learning across domains, a specific type of data heterogeneity can arise in horizontal FL or SL settings with partially overlapping feature sets. This problem arises from model sharing and the absence of some features at different sites, rather than merely missing values in the raw data. Park et al.\ \cite{park2024federated} proposed a model for partially overlapping clinical data using random forests. Their proposed framework involves each site training a random forest on its local data but sharing only the subset of trees built exclusively from the globally overlapping features. While this method allows each site to keep and utilize its locally trained trees that include private features, it completely discards these trees as part of the collaboration, potentially losing valuable information.

In this paper, we tackle the horizontal SL problem with partially overlapping features by developing inference-time strategies that aim to leverage most of the knowledge within the shared forest, including trees built with features unavailable locally. We propose and evaluate several methods designed to navigate these feature discrepancies, with the aim of maximizing the use of potentially informative trees. 

\section{Methods}
\label{methods}

\subsection{Problem Description}
\label{problem-description}


Imagine two hospitals A and B that have patient data related to diabetes. Due to different procedures for data generation and collection, they have different features related to patients. For instance, hospital A has records of pregnancies, blood pressure, skin thickness, BMI, age, and glucose, and hospital B has records of pregnancies, blood pressure, skin thickness, BMI, age, and insulin. While most features overlap, there are two features, glucose and insulin, that do not match in the two datasets. If these two hospitals decide to collaboratively train an RF model for diabetes prediction, they need to address the issue of missing features beforehand. 


In order to collaboratively train a global model, the two hospitals need to train local models on their private data. Then they share their model parameters to build a global model that has information from both datasets. For random forests, the learning process happens in one iteration, and the peers need to send their model parameters (trees) to other peers only once. To merge the models, the trees can be combined together to create a larger forest. In this way, for inference, all the trees that have been trained by different peers will cast a vote, and the majority class will be specified as the prediction.

Now imagine that the two hospitals, A and B, train local random forests on their private data. To create a global model, they share their trees and create a larger forest of trees, which is aggregated in both sites. The problem that occurs here is that during prediction, a sample will traverse all the trees and eventually will encounter a node in a tree that splits on a feature not available in the feature set of this sample. 

The simplest way to solve this issue is to omit the non-overlapping features from both sites before training. In our example, the two hospitals would remove the glucose and insulin features from the data and then train the models. In this way, they will train only on the intersection of their features, and the aforementioned problem will not occur. The downside of this simple solution, however, is that by taking the intersection of features, important information from the non-overlapping features will be lost. Here, we propose several methods to tackle the missing feature issue without losing the features completely, and then we compare these methods to the Intersection method as a baseline.

\subsection{Proposed Methods}
\label{proposed-methods}

\subsubsection{Feasible-Path Voting (FPV)}
\label{FPV}

As a first alternative to the Intersection method described previously, we introduce Feasible-Path Voting (FPV), a dynamic mechanism for inference. For each test sample, FPV constructs a unique subforest from the aggregated model. During this process, a tree is excluded from the voting ensemble if the sample's specific decision path (i.e., the traversal from the root to a leaf node) encounters a split conditional on a feature unavailable at the local site. This approach retains all trees in which the sample's decision path relies exclusively on its available features, even if those trees contain splits on non-overlapping features elsewhere. 

More formally, let $\mathcal{T}=\{\tau_1,\ldots,\tau_M\}$ be the pooled forest. For a test sample $\mathbf{x}$, with an available feature set $F_{\mathbf{x}}$, let $\mathcal{F}(\tau,\mathbf{x})$ represent the set of features inspected as $\mathbf{x}$ traverses a tree $\tau\in\mathcal{T}$. We define the subforest with a feasible path for sample $\mathbf{x}$ at each site, denoted $\mathcal{T}_{\mathrm{FPV}}(\mathbf{x})$, as the collection of all trees where this traversal only requires available features:

\begin{equation}
\mathcal{T}_{\mathrm{FPV}}(\mathbf{x}) \;=\; \bigl\{\,\tau\in\mathcal{T} \;:\; \mathcal{F}(\tau,\mathbf{x})\subseteq F_{\mathbf{x}} \,\bigr\}.
\end{equation}

The final prediction, $\hat{y}(\mathbf{x})$, is determined by a majority vote across this dynamically constructed subforest:

\begin{equation}
\hat{y}(\mathbf{x}) \;=\; \arg\max_{c\in\mathcal{Y}} \; \sum_{\tau\in\mathcal{T}_{\mathrm{FPV}}(\mathbf{x})} \mathbf{1}\{\tau(\mathbf{x})=c\},
\end{equation}
where $\tau(\mathbf{x})$ denotes the prediction of tree $\tau$ for
sample $\mathbf{x}$ under FPV.

\subsubsection{Probabilistic Routing}
\label{probabilistic-routing}
While Feasible-Path Voting (FPV) effectively utilizes trees whose decision paths are unaffected by missing features, it discards any tree where an unavailable feature is directly encountered. This discarding of trees may lead again to the loss of valuable information encoded in them. As an alternative to FPV, we propose Probabilistic Routing (PR), a method designed to leverage these otherwise excluded trees.

The core principle of PR is to resolve the traversal path stochastically when a deterministic decision cannot be made. During inference, when a test sample $\mathbf{x}$ encounters a decision node that splits on an unavailable feature, the tree is not immediately discarded. Instead, the subsequent path is determined probabilistically based on the empirical distribution of the training data at that node. Specifically, the probability of traversing to the left or right child is calculated as the fraction of training samples that followed each respective path. The sample is then routed to a child node by sampling from this probability distribution. This process is repeated for any subsequent unavailable features encountered until a leaf node is reached and a prediction is made.

A potential drawback of this approach is the introduction of noise, particularly if a probabilistic decision is made early in the tree's traversal (i.e., at a shallow depth). To mitigate this effect, we introduce a depth threshold, $d_{\text{thresh}}$. If a missing feature is encountered at a depth $d < d_{\text{thresh}}$, the tree is discarded as in FPV. However, if the encounter occurs at $d \ge d_{\text{thresh}}$, PR is used. This approach balances the benefit of retaining more trees against the risk of accumulating stochastic error.

To formalize this, let $n$ be a non-leaf node in a tree $\tau$. We first define the empirical traversal probabilities at that node based on the number of training samples, $N_n$, that reached it:

\begin{equation}
    p_n^L = \frac{N_n^L}{N_n} \quad \text{and} \quad p_n^R = \frac{N_n^R}{N_n},
\end{equation}
where $N_n^L$ and $N_n^R$ are the number of samples that proceeded to the left and right children, respectively.




The decision, then, to traverse to the right child ($n_R$) or the left child ($n_L$) is modeled as a Bernoulli trial:

\begin{equation}
    Z_n \sim \mathrm{Bernoulli}(p_n^R),
\end{equation}
where a value of $Z_n = 1$ directs the sample to the right child and $Z_n = 0$ to the left.

Let $\mathcal{F}_{<d_{\mathrm{thresh}}}(\tau,\mathbf{x})$ represent the set of features
inspected on the decision path of sample $\mathbf{x}$ in tree $\tau\in\mathcal{T}$ at
depths less than $d_{\text{thresh}}$. Then, the subforest for each sample is defined as:

\begin{equation}
\mathcal{T}_{\mathrm{PR}}(\mathbf{x}) = \{\tau \in \mathcal{T} \mid \mathcal{F}_{<d_{\mathrm{thresh}}}(\tau,\mathbf{x}) \subseteq F_{\mathbf{x}} \}.
\end{equation}

The final prediction is the majority vote over this dynamically assembled subforest:

\begin{equation}
    \hat{y}(\mathbf{x}) = \arg\max_{c \in \mathcal{Y}} \sum_{\tau \in \mathcal{T}_{\mathrm{PR}}(\mathbf{x})} \mathbf{1}\{\tau(\mathbf{x}) = c\},
\end{equation}
where $\tau(\mathbf{x})$ denotes the prediction of tree $\tau$ for sample $\mathbf{x}$ under PR with depth threshold $d_{\text{thresh}}$.

\subsubsection{Model Imputation}
\label{model-imputation}



Model Imputation (MI) offers an alternative to FPV and PR for navigating splits on unavailable features. Instead of immediately skipping a tree or choosing a path stochastically, MI resolves the split by imputing a value for the missing feature in real-time. This is achieved through training and sharing specialized imputation models for the private (i.e., non-overlapping) features at each site. 

The process consists of three main phases:

\begin{enumerate}
  \item \textbf{Training and Evaluation:} At each site, for every private feature, an imputation model is created that learns to predict that feature using the shared features as input. The performance of each imputer is then calculated using an evaluation metric. Specifically, for categorical private features, a Logistic Regression model is trained, and its quality is assessed via the mean $F1$-score over 3-fold cross-validation. For numerical features, a Linear Regression model is used, with its quality measured by the mean $R^2$ score, also over 3-fold cross-validation.
  \item \textbf{Sharing and Selection:} The trained imputation models, along with their corresponding quality scores, are shared among all peers. Upon receipt, each site stores the best-performing imputers for every non-local feature it might encounter during inference. For each feature, the imputer with the highest quality score is selected. If the scores are equal, the model trained on larger data is preferred.
  \item \textbf{Inference:} During inference, when a test sample $\mathbf{x}$ traverses a tree and encounters a split on a missing feature, the site uses its imputer to predict the value of the feature. This imputed value is then compared with the threshold at the node to decide whether to traverse through the right child or the left child. The same process is repeated for any missing feature until it reaches a leaf.
\end{enumerate}

Let $\mathbf{x}_O$ be the sub-vector in sample $\mathbf{x}$ corresponding to the values of the shared features $F_O$. During inference, if a tree traversal requires an unavailable feature $f$, its value is imputed by a machine learning model.

For a numerical feature $f$, its value $\hat{x}_f$ is imputed using a linear regression model: 

\begin{equation}
\hat{x}_f = \beta_{0,f} + \sum_{j \in F_O} \beta_{j,f} x_j,
\end{equation}
where $\beta_{j,f}$ are the model coefficients and the $x_j$ are the individual components of $\mathbf{x}_O$.

For a categorical feature $f$ with $K$ classes ${c_1, \dots, c_K}$, a Multinomial Logistic Regression model is used. The model first computes a linear score $z_k$ for each class $k$:

\begin{equation}
z_k = \beta_{0,k} + \sum_{j \in F_O} \beta_{j,k} x_j,
\end{equation}
where $\beta_{j,k}$ are the model coefficients for class $k$. These scores are then converted into probabilities using the Softmax function:

\begin{equation}
P(f=c_k | \mathbf{x}_O) = \frac{\exp(z_k)}{\sum_{i=1}^{K} \exp(z_i)}.
\end{equation}

The imputed value $\hat{x}_f$ is then the class with the highest probability:

\begin{equation}
\hat{x}_f = \underset{c_k \in \{c_1, \dots, c_K\}}{\arg\max} \; P(f=c_k | \mathbf{x}_O).
\end{equation}

Let $\tau(\mathbf{x})$ be the prediction resulting from a tree traversal
that applies the imputation rule for any unavailable features. For a given sample
$\mathbf{x}$, let $\mathcal{T}_{\mathrm{MI}}(\mathbf{x}) \subseteq \mathcal{T}$ denote
the set of trees for which all required imputations can be performed. The final prediction
for the sample is then the majority vote over this subforest:

\begin{equation}
\hat{y}(\mathbf{x}) \;=\; \arg\max_{c\in\mathcal{Y}} \; \sum_{\tau\in\mathcal{T}_{\mathrm{MI}}(\mathbf{x})} \mathbf{1}\{\tau(\mathbf{x})=c\}.
\end{equation}

In this paper, we do not evaluate MI as a standalone method. Rather, we adopt the concept for other methods discussed in the following.

\subsubsection{Informed Probabilistic Routing}
\label{informed-probabilistic-routing}
While the PR method navigates splits on unavailable features stochastically, its routing is based only on the distribution of training data at a node, resulting in fixed probabilities that are independent of the test sample's available feature values. Informed Probabilistic Routing (IPR) aims to create a more data-driven choice by incorporating sample-specific information. IPR leverages the imputation models developed for MI, but uses their output to guide a probabilistic decision rather than a deterministic one.

As in MI, each site trains and shares imputation models for its private features. The standard deviation for the numerical features is also computed and shared. During inference, if a sample encounters a node that splits on an unavailable numerical feature, an imputed value is generated for that feature. However, instead of making a hard decision based on this value (as in MI), IPR uses the normalized distance between the imputed value and the node's threshold to define a probability for traversing left or right. Intuitively, a large distance between the imputed value and the threshold allows for a more confident routing decision (towards the left or the right), while a small distance reflects low confidence. We capture this by applying the sigmoid function to the normalized distance: a large positive distance (a confident right guess) is mapped to a probability close to $1$, a large negative distance (a confident left guess) to a probability close to $0$, and a distance near zero to a probability close to $0.5$.

If the missing feature is categorical, then instead of using the imputed value, we get the probability distributions from the Logistic Regression model. Then, we can calculate the probabilities of traversing through the left or the right child by summing the predicted probabilities of all the categories that belong to either path.

Suppose the traversal for a sample $\mathbf{x}$ with a set of features $F_{\mathbf{x}}$ reaches a node $n$, which splits on an unavailable numerical feature $f \notin F_\mathbf{x}$ with threshold $t_n$. Let $\hat{x}_{f}$ be the value for this feature imputed by the selected model $I_{f}^\star$:

\begin{equation}
\hat{x}_{f} = I_{f}^\star(\mathbf{x}_O).
\end{equation}

The normalized distance, $d_n$, between the imputed value and the threshold is calculated as:

\begin{equation}
d_n = \frac{\hat{x}_{f}-t_n}{\sigma_{f}},
\end{equation}
where $\sigma_f$ is the global standard deviation for feature $f$ averaged across all sites that contain that feature. The probability of traversing to the right child is then computed using the Sigmoid function:

\begin{subequations}
\begin{align}
p_{\mathrm{R}}(n \mid \hat{x}_{f}) &= \frac{1}{1 + e^{-d_n}}, \label{eq:routing_R}\\
p_{\mathrm{L}}(n \mid \hat{x}_{f}) &= 1 - p_{\mathrm{R}}(n \mid \hat{x}_{f}).
\end{align}
\end{subequations}

If the missing feature $f$ at node $n$ is categorical, then instead of imputing a value, the selected imputer $I_{f}^\star$ produces the probability vector for all the classes of feature $f$. Let the set of possible classes for $f$ be $C_f = \{c_1, \dots, c_K\}$. Then, using the sample's shared features $\mathbf{x}_O$, the imputer produces a vector of probabilities $\mathbf{\hat{p}}$ for the classes of the unavailable feature $f$:

\begin{equation}
\hat{\mathbf{p}}_f = (\hat{p}_1, \dots, \hat{p}_K) \quad \text{where} \quad \hat{p}_k = P(f=c_k | \mathbf{x}_O).
\end{equation}

The probability of traversing to the left child is then the sum of the probabilities of all classes $c_k$ that satisfy the node's split condition $c_k \le t_n$:

\begin{subequations}\label{eq:cat_routing}
\begin{align}
p_{\mathrm{L}}(n \mid \hat{\mathbf{p}}_f) &= \sum_{k : c_k \le t_n} \hat{p}_k, \label{eq:cat_routing_L}\\
p_{\mathrm{R}}(n \mid \hat{\mathbf{p}}_f) &= 1 - p_{\mathrm{L}}(n \mid \hat{\mathbf{p}}_f). \label{eq:cat_routing_R}
\end{align}
\end{subequations}

The sample then proceeds to the right child with probability $p_{\text{R}}$ and to the left child with probability $p_{\text{L}}$. If the imputation part fails, IPR falls back to the PR rule at that node, using the empirical left and right proportions based on the training data.

With $\tau(\mathbf{x})$ representing the prediction from a tree using this method, the final prediction for sample $\mathbf{x}$ is the majority vote:

\begin{equation}
\hat{y}(\mathbf{x}) \;=\; \arg\max_{c\in\mathcal{Y}} \;
\sum_{\tau\in\mathcal{T}_{\mathrm{IPR}}(\mathbf{x})} \mathbf{1}\{\tau(\mathbf{x})=c\},
\end{equation}
where $\mathcal{T}_{\mathrm{IPR}}(\mathbf{x}) \subseteq \mathcal{T}$ represents the set of trees that are not discarded by the fallback mechanism.

\subsubsection{Marginal Prediction}
\label{marginal-prediction}

In previous sections, we explored methods that resolved a split on an unavailable feature by selecting a single path through the left or the right child, stochastically or deterministically. Here, we propose the Marginal Prediction (MP) method built upon IPR, in which, rather than choosing a single branch at a split on an unavailable feature, we explore both possible paths. MP marginalizes over the missing split by recursively calculating a weighted average of the class probability distributions from the entire left and right subtrees. This idea is conceptually related to the weighting strategy used in C4.5 for handling missing values, where instances are distributed fractionally across branches. However, here the trees are built in standard random forests without fractional sample assignments, and MP is applied only at inference time using probabilities derived from imputation models to weight the branches.

Let $\mathbf{p}_{\text{leaf}}(n)$ be the class distribution at leaf node $n$:

\begin{equation}
\mathbf{p}_{\text{leaf}}(n)
\;=\;
\frac{\mathbf{c}(n)}{\sum_{y\in\mathcal{Y}} [\mathbf{c}(n)]_y},
\end{equation}
where $\mathbf{c}(n)$ are the class counts stored at node $n$. Let \(\mathcal{T}=\{\tau_1,\ldots,\tau_M\}\) be the pooled forest. For tree \(\tau\) and node \(n\) with children \(n_L\) and \(n_R\), split feature \(f_n\), and threshold \(t_n\), we define the class probability distribution vector at \(n\) for sample \(\mathbf{x}\) as \(\mathbf{p}_\tau(n\,|\,\mathbf{x})\), given by:

\begin{equation}
\footnotesize
\mathbf{p}_\tau(n \mid \mathbf{x}) =
\begin{cases}
  \mathbf{p}_{\text{leaf}}(n), & \text{$n$ is a leaf node}, \\
  \mathbf{p}_\tau(n_L \mid \mathbf{x}), & \text{$f_n \in F_{\mathbf{x}}, x_{f_n} \le t_n$}, \\
  \mathbf{p}_\tau(n_R \mid \mathbf{x}), & \text{$f_n \in F_{\mathbf{x}}, x_{f_n} > t_n$}, \\[3pt]
  \makebox[0pt][l]{$p_{\mathrm L}(n \mid \mathbf{x}) \mathbf{p}_\tau(n_L \mid \mathbf{x})$} & \\[-2pt]
  \quad +\, p_{\mathrm R}(n \mid \mathbf{x}) \mathbf{p}_\tau(n_R \mid \mathbf{x}), & \text{$f_n \notin F_{\mathbf{x}}$}.
\end{cases}
\end{equation}

Here, $p_{\mathrm{L}}(n\mid \mathbf{x})$ and $p_{\mathrm{R}}(n\mid \mathbf{x})$ are the probabilities for traversing through the left child and the right child, calculated by the IPR method.

The final class probability vector for a single tree is obtained by calculating $\mathbf{p}_\tau(\text{root} \mid \mathbf{x})$. The final predicted class $\hat{y}(\mathbf{x})$ is then the class with the highest probability in the average of these vectors from all trees in the forest $\mathcal{T}$:

\begin{equation}
\hat{y}(\mathbf{x})
\;=\;
\underset{c \in \mathcal{Y}}{\arg\max}\;
\left[\, \frac{1}{|\mathcal{T}|} \sum_{\tau \in \mathcal{T}} \mathbf{p}_\tau(\text{root} \mid \mathbf{x}) \,\right]_c .
\end{equation}

\subsubsection{Surrogate Splits}
\label{surrogate-splits}

The final method proposed in this paper to resolve the missing feature problem is the classic surrogate splits introduced in CART \cite{breiman2017classification}. This method was originally introduced to compensate for the missing values. The main idea is to find an alternate decision rule that can best mimic the partitioning of the data achieved by the primary split. Here, we utilize this idea to find a surrogate from the overlapping features for any split that relies on a private feature. This allows trees trained on one site's private data to be interpretable at other sites. 

After training the local RF models, each site scans all the nodes that split on a private feature. For each node $n$ in tree $\tau$ that splits on a private feature $f_p$, the goal is to find a feature $f_o \in F_O$ that most accurately mimics the partitioning of the primary split. This is done by testing all the possible thresholds for all the overlapping features to find the best feature-threshold set as a surrogate for the original. The quality of this surrogate is assessed by calculating the adjusted agreement $\operatorname{adj}(f_p, f_o)$, which is maximized when the surrogate sends the same samples to the left or right child as the primary split. The adjusted agreement is calculated as follows:

\begin{enumerate}
  \item \textbf{Raw Agreement:} First we calculate the raw agreement $n_{\text{surr}}$, which is the number of training samples at node $n$ that are sent to the same child node by both the primary split and the candidate surrogate split. That is, the metric counts the specific samples assigned to the same partition, not simply the total counts in each child node.
  \item \textbf{Adjusted Agreement:} Since raw agreement can be misleading, we need to adjust it to account for chance agreement. For instance, suppose that a primary split at a node with 100 samples sends 95 samples to the left child and only 5 samples to the right child. A surrogate that defaults to the majority rule (i.e., sending all the samples to the left) would achieve a raw agreement of 95 (95\% in percentage). This high score is misleading because the surrogate has not learned to identify the 5 samples that should go to the right. To account for this, we calculate the adjusted agreement: 

\begin{equation}
\operatorname{adj}(f_p, f_o)
=
\frac{n_{\text{surr}} - n_{\text{maj}}}{n_{\text{total}} - n_{\text{maj}}},
\end{equation}
where $n_{\text{surr}}$ is the raw agreement, $n_{\text{maj}}$ is the number of correct assignments when all samples are assigned to the child node with the majority of samples, and $n_{\text{total}}$ is the total number of samples that reached the node.

\end{enumerate}

The surrogates are then attached as attributes for the tree and can be used at other sites. During inference, if the test sample $\mathbf{x}$ reaches a node with an unavailable feature, it uses the chosen surrogate (with threshold $t_{\text{surr}}$) to resolve that split and continue traversal. If no surrogate is available, for example because no candidate achieved a satisfactory adjusted agreement, the method falls back to the PR rule at that node, which itself may discard shallow trees as in FPV. Let $\mathcal{T}_{\mathrm{SS}}(\mathbf{x}) \subseteq \mathcal{T}$ be the subset of trees that are not discarded during this process. The final prediction $\hat{y}(\mathbf{x})$ is the majority vote from this subset, where $\tau(\mathbf{x})$ is the prediction of a single tree:

\begin{equation}
\hat{y}(\mathbf{x})
=
\underset{c \in \mathcal{Y}}{\arg\max} \;
\sum_{\tau \in \mathcal{T}_{\mathrm{SS}}(\mathbf{x})}
\mathbf{1}\{\tau(\mathbf{x}) = c\}.
\end{equation}

\section{Experiments Setup}
\label{experiments-setup}

To evaluate the performance of the proposed methods, we have conducted extensive simulations, testing multiple datasets across different scenarios. 

\subsection{Simulation Environment}
\label{simulation-environment}
In order to test our methods, we simulated an SL network in which several peers collaboratively train RF models in a cross-silo setting. They form a fully connected network in which they share their parameters directly with other peers without a central server. The process starts by having each site train a local RF model on its private data. Following local training, they share their parameters (i.e., model estimators and method-specific parameters) with other peers. Upon receiving the parameters from other peers, each site aggregates them with its own parameters to create the final global model. For RFs, this is achieved by aggregating the trees to form a larger forest, which will be used for prediction. 

We designed the experiments across a total of 16 distinct scenarios, varying both the number of participating peers and the degree of feature overlap. More specifically, we simulated networks of 2, 3, 4, and 6 peers. For each network size, we considered eight levels of feature heterogeneity, with target Jaccard similarities ranging from 0.2 to 0.9. 






To mitigate bias from sample and feature assignment, each experiment was repeated for 100 iterations. In each iteration, data was distributed evenly and randomly among the peers. Then, at each peer, the data was divided into a train set and a test set with a ratio of 75\% and 25\%, respectively. For feature assignment, we first specified a target pairwise Jaccard similarity for the feature sets in the given scenario and, based on this target, computed the number of features that should be shared across all peers. Let $F$ denote the total number of features, $N$ the number of peers, and $O$ the number of overlapping features. With $P$ private features per peer, we have $F = O + N P$ and Jaccard similarity:

\begin{equation}
J \;=\; \frac{O}{O + 2P}.
\end{equation}

Solving for $O$ gives:

\begin{equation}
O \;=\; \frac{F}{1 + N \left(\frac{1}{2J} - \frac{1}{2}\right)},
\end{equation}
which we round to the nearest integer and use as the size of the overlapping feature set. A random subset of this size was selected as the overlapping feature set, and the remaining features were treated as private. These private features were then distributed across peers in a balanced way, ensuring that the resulting feature sets approximately achieve the desired Jaccard overlap.

\subsection{Datasets}
\label{datasets}

Our experiments were conducted on a total of nine datasets, comprising six publicly available, real-world datasets and three synthetic datasets. The real-world datasets cover a range of medical and public-health applications and differ in sample size, number and type of features, and class imbalance. Specifically, we use the Glioma dataset \cite{tasci2022hierarchical} for identifying glioma brain tumors, the Thyroid dataset \cite{borzooei2024machine} for predicting recurrence of well-differentiated thyroid cancer, the Pima Indians Diabetes dataset \cite{smith1988using} for predicting diabetes, the Gallstone dataset \cite{esen2024early} for predicting gallstone disease, and two datasets derived from the CDC Diabetes Health Indicators data \cite{burrows2017incidence}. For the binary version (CDC-2Class), we drew a stratified random sample of 1,000 instances from the CDC binary variant. For the three-class version (CDC-3Class), we constructed a balanced random sample of 1,000 instances from the original CDC three-class dataset. The three synthetic datasets, denoted S-500-Num, S-500-Cat, and S-500-Mix, contain 16 numerical features, 16 categorical features, and an equal mix of both (8 numerical and 8 categorical), respectively. All features were generated to contribute evenly to a binary target. This allows us to study the behavior of the methods in a setting where the feature space is well-controlled. Table~\ref{tab:dataset_specs} summarizes the key characteristics of all datasets used in the experiments.

\begin{table}[!t]
\caption{Summary of dataset specifications used in the experiments. The imbalance ratio is defined as the ratio between the size of the majority class and the size of the minority class (or the smallest class in the multiclass case).}
\label{tab:dataset_specs}
\centering
\scriptsize
\setlength{\tabcolsep}{5.5pt}
\begin{tabular}{|l|c|c|c|c|c|}
\hline
\textbf{Dataset} &
\textbf{Samples} &
\textbf{\begin{tabular}[c]{@{}c@{}}Numerical\\ Features\end{tabular}} &
\textbf{\begin{tabular}[c]{@{}c@{}}Categorical\\ Features\end{tabular}} &
\textbf{Classes} &
\textbf{\begin{tabular}[c]{@{}c@{}}Imbalance\\ Ratio\end{tabular}} \\ \hline \hline

Glioma              & 839   & 1  & 22 & 2 & 1.38 \\ \hline
Thyroid             & 383   & 1  & 14 & 2 & 2.55 \\ \hline
Diabetes            & 768   & 8  & 0  & 2 & 1.87 \\ \hline
Gallstone           & 319   & 31 & 7  & 2 & 1.02 \\ \hline
CDC-3Class          & 1,000  & 3  & 18 & 3 & 1.00 \\ \hline
CDC-2Class          & 1,000  & 3  & 18 & 2 & 1.00 \\ \hline
S-500-Num          & 500  & 16 & 0  & 2 & 1.15 \\ \hline
S-500-Cat          & 500  & 0  & 16 & 2 & 1.00 \\ \hline
S-500-Mix          & 500  & 8  & 8  & 2 & 1.02 \\ \hline

\end{tabular}
\end{table}

\subsection{Evaluation Metrics}
\label{evaluations-metrics}
To evaluate our methods, we selected three standard metrics suitable for classification tasks: the Area Under the Precision--Recall Curve (AUPRC), the Area Under the Receiver Operating Characteristic Curve (AUC), and the Matthews Correlation Coefficient (MCC). AUPRC, which ranges from 0 to 1, summarizes the trade-off between precision and recall across decision thresholds and is particularly informative on imbalanced datasets. AUC, also ranging from 0 to 1, measures the model's ability to distinguish between classes across all decision thresholds. Finally, MCC, which ranges from $-1$ to $+1$, is a robust metric that produces a high score only if the classifier performs well on both majority and minority classes. For multiclass problems, AUPRC and AUC are computed in a one-vs-rest manner and averaged across classes.

\section{Results}
\label{results}

In this section we present and discuss the results of our experiments. As described above, each dataset was evaluated across 32 distinct scenarios, defined by the combination of four peer configurations (2, 3, 4, and 6 peers) and eight Jaccard overlap levels. The reported scores correspond to the mean performance in the system, averaged across peers and over 100 random repetitions. Here, we report detailed AUPRC results for the S-500-Mix and CDC-2Class datasets. Results for the remaining datasets and additional metrics (AUC, MCC) are provided in the Supplementary Material.

\begin{figure}[!t]
\centering
\includegraphics[width=\columnwidth]{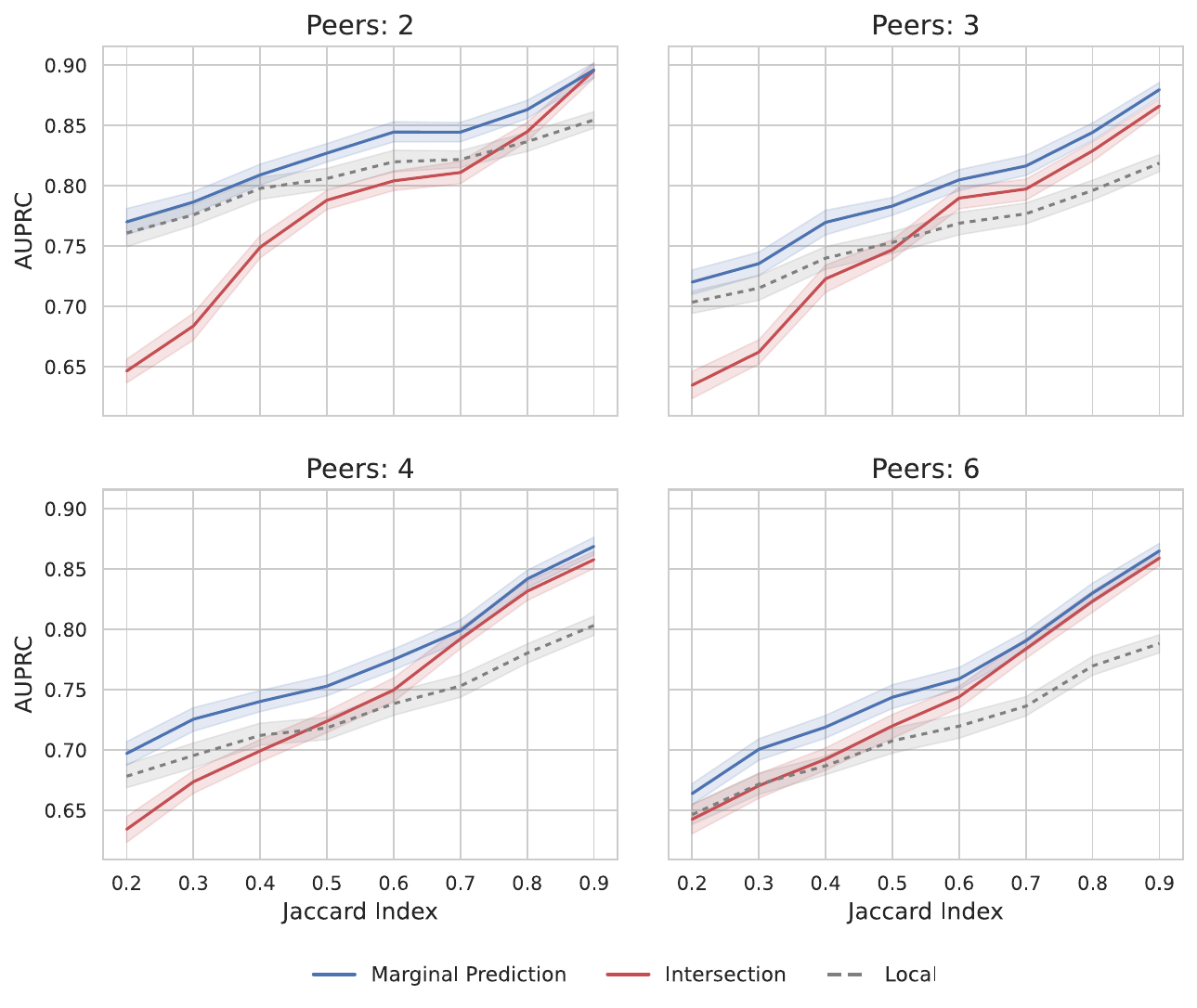}
\caption{AUPRC scores for the S-500-Mix dataset across different peer scenarios and overlap levels, comparing Marginal Prediction with Local and Intersection.}
\label{fig-auprc-lines-s500mix}
\end{figure}

Figures~\ref{fig-auprc-lines-cdc} and~\ref{fig-auprc-lines-s500mix} show the AUPRC scores for CDC-2Class and S-500-Mix, respectively, comparing the MP method, as the overall best-performing method, with Intersection and Local training across four peer configurations and eight feature-overlap levels. The curves illustrate that performance increases with higher overlap and that MP consistently outperforms both Intersection and Local across all scenarios. Line plots for the remaining datasets and metrics are provided in the Supplementary Material.

\begin{figure}[!t]
\centering
\includegraphics[width=\columnwidth]{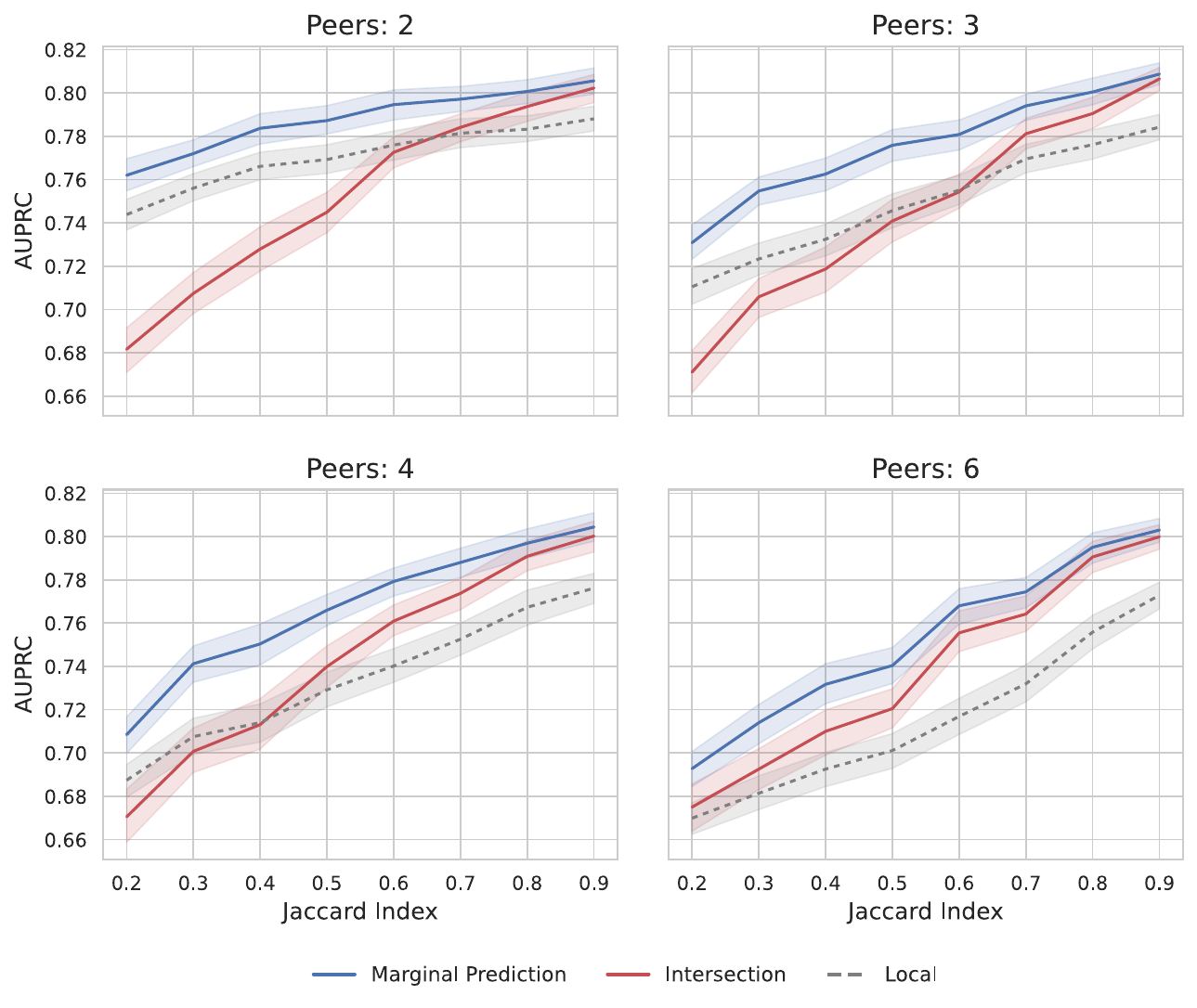}
\caption{AUPRC scores for the CDC-2Class dataset across different peer scenarios and overlap levels, comparing Marginal Prediction with Local and Intersection.}
\label{fig-auprc-lines-cdc}
\end{figure}

\begin{figure}[!t]
\centering
\includegraphics[width=\columnwidth]{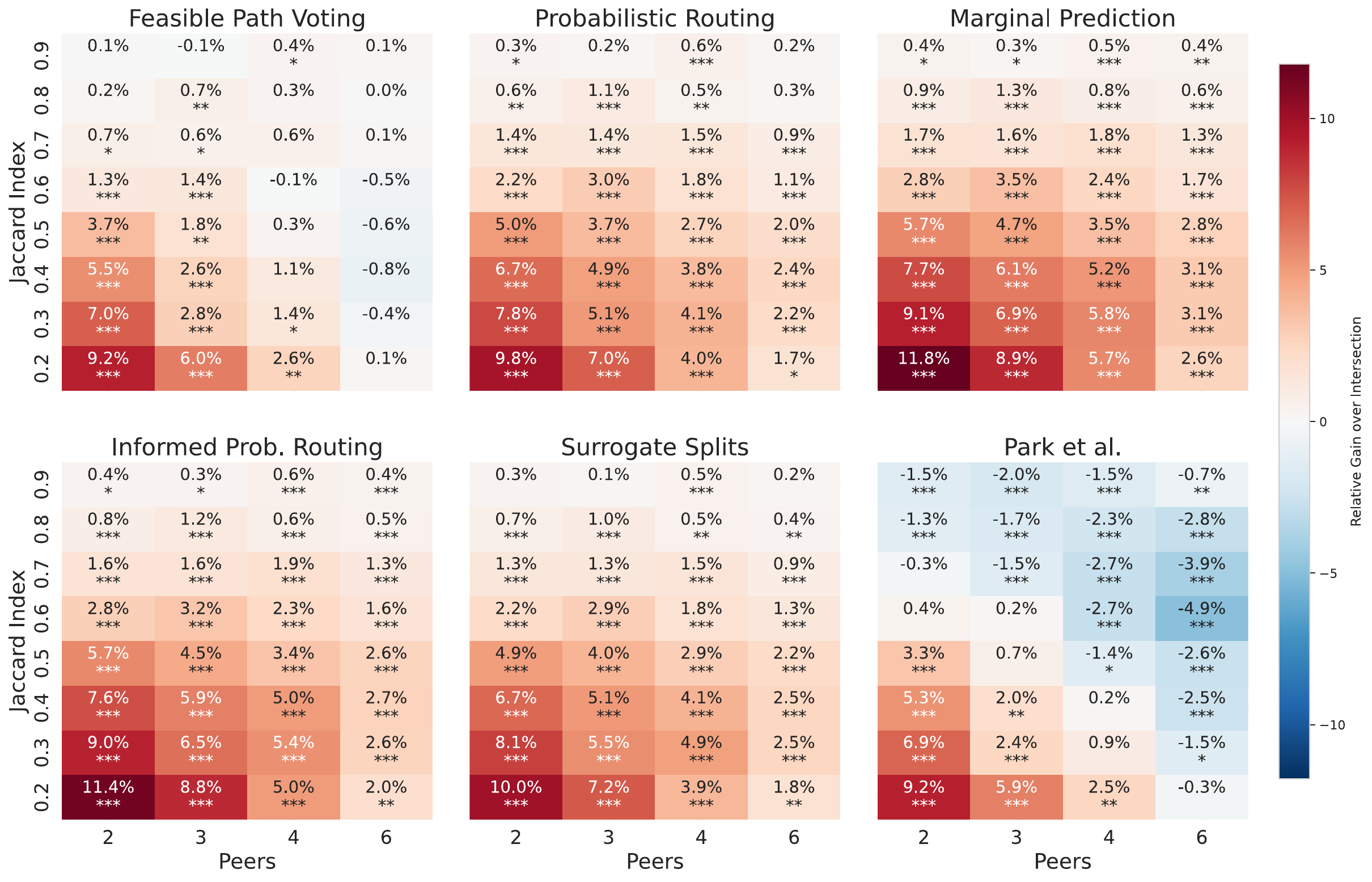}
\caption{Relative gain in AUPRC for each method compared to Intersection for the CDC-2Class dataset across different peer scenarios and overlap levels.}
\label{fig-auprc-heatmap-cdc}
\end{figure}

\begin{figure}[!t]
\centering
\includegraphics[width=\columnwidth]{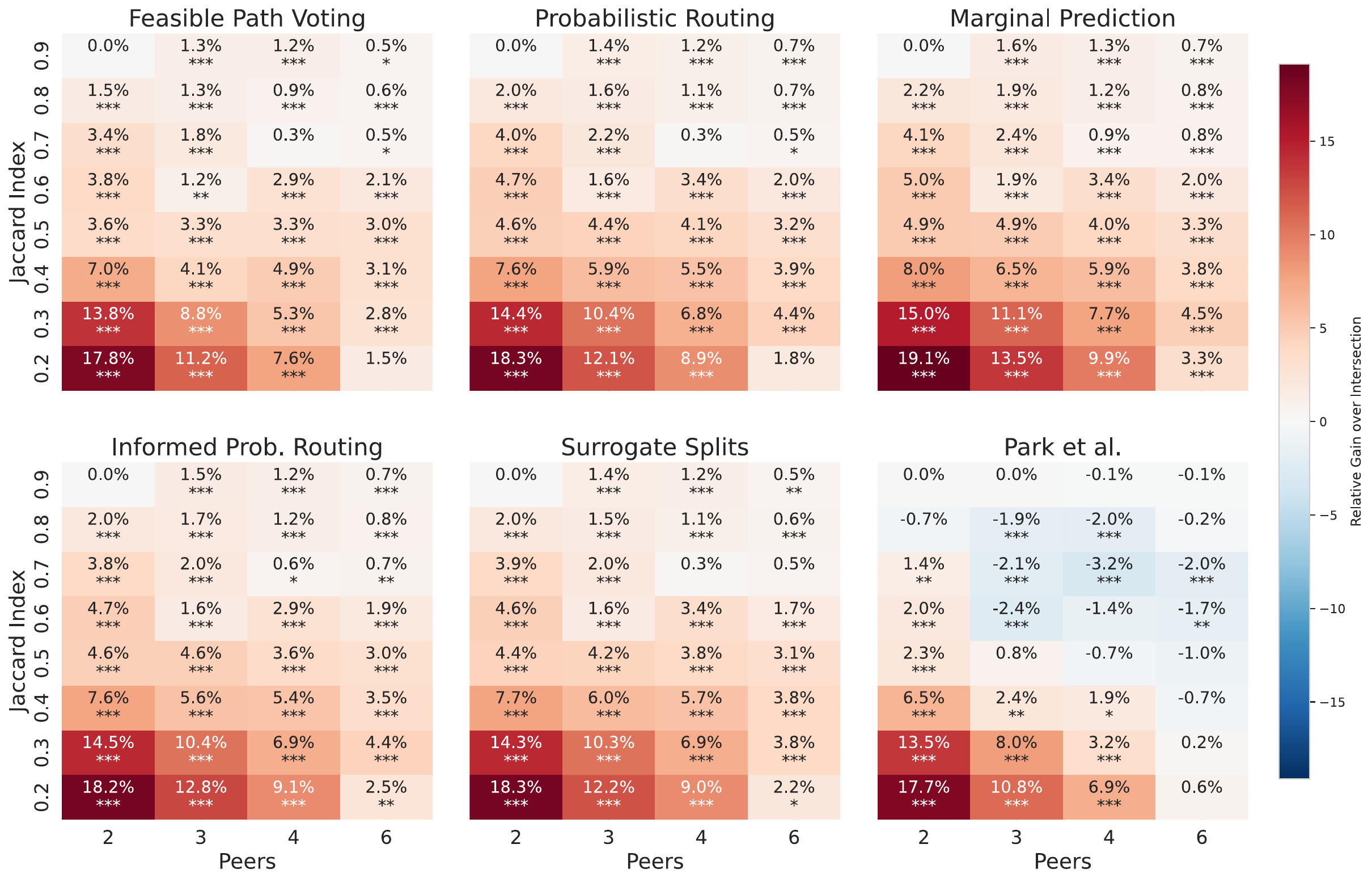}
\caption{Relative gain in AUPRC for each method compared to Intersection for the S-500-Mix dataset across different peer scenarios and overlap levels.}
\label{fig-auprc-heatmap-s500mix}
\end{figure}

To quantify the gain in performance of each method over the baseline, Figures~\ref{fig-auprc-heatmap-cdc} and~\ref{fig-auprc-heatmap-s500mix} show heatmaps of the relative improvement with respect to Intersection for CDC-2Class and S-500-Mix. For each method $m$ and scenario $s$ (a particular combination of peers and overlap level), we compute:

\begin{equation}
    G_{m,s} \;=\; \frac{\mathrm{Score}_{m,s} - \mathrm{Score}_{\text{Int},s}}{\mathrm{Score}_{\text{Int},s}} \,,
\end{equation}
where $\mathrm{Score}_{m,s}$ denotes the score for method $m$ in scenario $s$, and $\mathrm{Score}_{\text{Int},s}$ is the corresponding value for the Intersection baseline. These heatmaps demonstrate that our methods outperform the baseline in almost all scenarios, with the largest relative gains occurring at low overlap. In these settings, the proposed strategies are capable of utilizing information from non-overlapping features that are entirely discarded by Intersection. Heatmaps for the remaining datasets and metrics are provided in the Supplementary Material.

Tables~\ref{tab-auprc-cdc-2class} and~\ref{tab-auprc-s500mix} report AUPRC scores for all methods on CDC-2Class and S-500-Mix, respectively. For each dataset, the tables show the mean AUPRC for every peer configuration at two representative overlap levels, Jaccard $J=0.3$ (low overlap) and $J=0.7$ (high overlap). The results show that the proposed methods outperform the approach of Park et al.\ as well as Local and Intersection. Full tables covering all overlap levels, additional datasets, and further metrics are provided in the Supplementary Material.

\begin{table*}[!t]
\caption{AUPRC for the CDC-2Class dataset across peer scenarios for two representative overlap levels ($J=0.3$, $J=0.7$).}
\label{tab-auprc-cdc-2class}
\centering
\footnotesize
\begin{tabular}{|c|c|cccccccc|}
\hline
\textbf{Peers} & \textbf{Jaccard} & \textbf{Local} & \textbf{Intersection} & \textbf{Park et al.} & \textbf{FPV} & \textbf{PR} & \textbf{IPR} & \textbf{MP} & \textbf{SS} \\ \hline \hline
\multirow{2}{*}{2} & 0.3 & 0.76 & 0.71 & 0.76 & 0.76 & 0.76 & \textbf{0.77} & \textbf{0.77} & 0.76 \\ \cline{2-10}
 & 0.7 & 0.78 & 0.78 & 0.78 & 0.79 & \textbf{0.80} & \textbf{0.80} & \textbf{0.80} & 0.79 \\ \hline
\multirow{2}{*}{3} & 0.3 & 0.72 & 0.71 & 0.72 & 0.73 & 0.74 & \textbf{0.75} & \textbf{0.75} & 0.74 \\ \cline{2-10}
 & 0.7 & 0.77 & 0.78 & 0.77 & \textbf{0.79} & \textbf{0.79} & \textbf{0.79} & \textbf{0.79} & \textbf{0.79} \\ \hline
\multirow{2}{*}{4} & 0.3 & 0.71 & 0.70 & 0.71 & 0.71 & 0.73 & \textbf{0.74} & \textbf{0.74} & 0.73 \\ \cline{2-10}
 & 0.7 & 0.75 & 0.77 & 0.75 & 0.78 & \textbf{0.79} & \textbf{0.79} & \textbf{0.79} & \textbf{0.79} \\ \hline
\multirow{2}{*}{6} & 0.3 & 0.68 & 0.69 & 0.68 & 0.69 & \textbf{0.71} & \textbf{0.71} & \textbf{0.71} & \textbf{0.71} \\ \cline{2-10}
 & 0.7 & 0.73 & 0.76 & 0.73 & \textbf{0.77} & \textbf{0.77} & \textbf{0.77} & \textbf{0.77} & \textbf{0.77} \\ \hline
\end{tabular}
\end{table*}

\begin{table*}[!t]
\caption{AUPRC for the S-500-Mix dataset across peer scenarios for two representative overlap levels ($J=0.3$, $J=0.7$).}
\label{tab-auprc-s500mix}
\centering
\footnotesize
\begin{tabular}{|c|c|cccccccc|}
\hline
\textbf{Peers} & \textbf{Jaccard} & \textbf{Local} & \textbf{Intersection} & \textbf{Park et al.} & \textbf{FPV} & \textbf{PR} & \textbf{IPR} & \textbf{MP} & \textbf{SS} \\ \hline \hline
\multirow{2}{*}{2} & 0.3 & 0.78 & 0.68 & 0.78 & 0.78 & 0.78 & 0.78 & \textbf{0.79} & 0.78 \\ \cline{2-10}
 & 0.7 & 0.82 & 0.81 & 0.82 & \textbf{0.84} & \textbf{0.84} & \textbf{0.84} & \textbf{0.84} & \textbf{0.84} \\ \hline
\multirow{2}{*}{3} & 0.3 & 0.72 & 0.66 & 0.72 & 0.72 & 0.73 & 0.73 & \textbf{0.74} & 0.73 \\ \cline{2-10}
 & 0.7 & 0.78 & 0.80 & 0.78 & 0.81 & 0.81 & 0.81 & \textbf{0.82} & 0.81 \\ \hline
\multirow{2}{*}{4} & 0.3 & 0.70 & 0.67 & 0.70 & 0.71 & 0.72 & 0.72 & \textbf{0.73} & 0.72 \\ \cline{2-10}
 & 0.7 & 0.75 & 0.79 & 0.77 & 0.79 & 0.79 & \textbf{0.80} & \textbf{0.80} & 0.79 \\ \hline
\multirow{2}{*}{6} & 0.3 & 0.67 & 0.67 & 0.67 & 0.69 & \textbf{0.70} & \textbf{0.70} & \textbf{0.70} & \textbf{0.70} \\ \cline{2-10}
 & 0.7 & 0.74 & 0.78 & 0.77 & \textbf{0.79} & \textbf{0.79} & \textbf{0.79} & \textbf{0.79} & \textbf{0.79} \\ \hline
\end{tabular}
\end{table*}

\begin{figure*}[!t]
\centering
\includegraphics[width=0.7\textwidth]{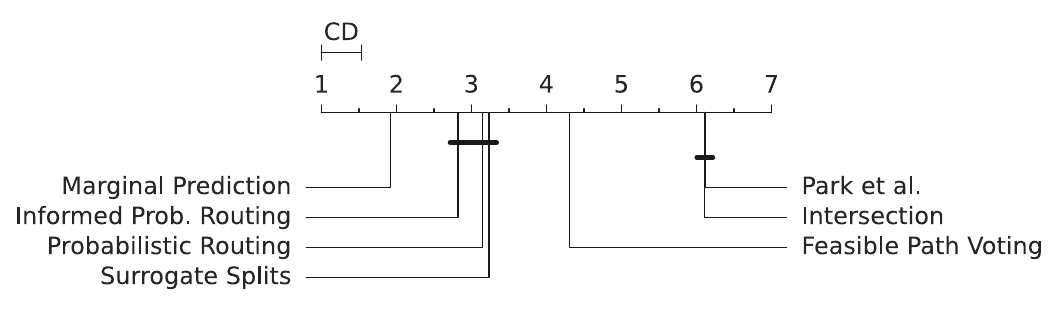}
\caption{Critical difference diagram of average ranks for AUPRC across all datasets, peer scenarios, and overlap levels.}
\label{fig-cd-global}
\end{figure*}

To obtain an overall comparison of the methods, we applied the Friedman/Nemenyi procedure to rank them across all datasets and scenarios. Figure~\ref{fig-cd-global} shows the resulting critical difference (CD) diagram for AUPRC. As the diagram shows, Marginal Prediction achieves the best average rank and is significantly better than all other methods, while Intersection and Park et al.\ rank last. IPR, PR, and SS occupy the next three positions but are not mutually significantly different, which suggests that their relative performance depends on the specific dataset and overlap setting. This highlights that the strength of a method can vary with properties such as the amount of feature overlap and the structure of the data. CD diagrams computed separately per dataset and per overlap level are provided in the Supplementary Material.

\section{Discussion}
\label{discussion}

In this paper, we introduced several inference-time strategies to address feature heterogeneity, and in particular the problem of partially overlapping feature sets in a Swarm Learning setting. These strategies include both probabilistic and deterministic methods that resolve splits on unavailable features in the global random forest during inference. Our results show that the proposed methods outperform locally trained models as well as the Intersection baseline. We considered a wide range of overlap scenarios, from very low to high Jaccard similarity among peers, to obtain a broad picture of how the methods behave in different settings. As demonstrated, absolute performance improves with higher overlap, since fewer splits rely on unavailable features and more decision paths can be followed normally. At the same time, for methods that rely on imputers or surrogate splits, the quality of these models also depends on the number of overlapping features, which explains the decrease in performance at very low overlap.

It is important to note that this work focuses on an SL network operating in a cross-silo setting rather than a cross-device setting. In cross-device scenarios, many (potentially tens, hundreds or even millions of) devices collaborate in a federated network to build a global model. In cross-silo scenarios, by contrast, only a small number of organizations (e.g., hospitals) participate. Since SL uses a fully connected peer-to-peer network, it is naturally suited to such cross-silo collaborations, which is the setting considered in this study.

Finally, although the idea behind FL and SL systems is to make privacy-preserving collaboration possible by having the private data remained at local sites, they are not inherently privacy-preserving: as the information shared through model updates or shared trees can still be exploited by malicious parties. In practice, such systems are typically safeguarded by additional mechanisms such as homomorphic encryption or differential privacy. In this paper, we assumed honest participants and focused on feature heterogeneity and inference-time handling of missing features. Integrating our methods with stronger privacy guarantees and robustness against adversarial behavior can be the focus of future research.

\section{Conclusion}
\label{conclusion}

In this paper, we studied feature heterogeneity in Swarm Learning with Random Forests, focusing on horizontal collaboration where sites have partially overlapping feature sets. Instead of restricting training to the shared feature intersection, we proposed several inference-time strategies that handle missing splits when a pooled tree tests a feature that is unavailable locally. The proposed approaches use deterministic or probabilistic routing to keep more trees usable during prediction. We evaluated the methods across multiple datasets and a wide range of scenarios, varying both the number of peers and the degree of feature overlap. Compared with the intersection baseline and with locally trained models, our methods achieved higher performance, showing that inference-time handling of missing split features can make decentralized Random Forest collaboration more effective under partial feature overlap.


\bibliographystyle{IEEEtran}
\bibliography{references}

\vfill

\end{document}